# Text Region Extraction from Business Card Images for Mobile Devices

A. F. Mollah[+], S. Basu*, N. Das*, R. Sarkar*, M. Nasipuri*, M. Kundu*

[+] School of Mobile Computing and Communication, Jadavpur University, India
* Department of Computer Science & Engineering, Jadavpur University, India

Email: afmollah@gmail.com

**Abstract**

Designing a Business Card Reader (BCR) for mobile devices is a challenge to the researchers because of huge deformation in acquired images, multiplicity in nature of the business cards and most importantly the computational constraints of the mobile devices. This paper presents a text extraction method designed in our work towards developing a BCR for mobile devices. At first, the background of a camera captured image is eliminated at a coarse level. Then, various rule based techniques are applied on the Connected Components (CC) to filter out the noises and picture regions. The CCs identified as text are then binarized using an adaptive but light-weight binarization technique. Experiments show that the text extraction accuracy is around 98% for a wide range of resolutions with varying computation time and memory requirements. The optimum performance is achieved for the images of resolution 1024x768 pixels with text extraction accuracy of 98.54% and, space and time requirements as 1.1 MB and 0.16 seconds respectively.

## 1. Introduction

The usage of business cards is not only limited to business groups but these are also extensively used by common people including teachers, doctors, lawyers, etc. So, managing the cards with an album does not satisfy those who have handheld mobile devices like cell phone, PDA, etc. An efficient management can be to have the required information populated from the cards into the mobile device with the help of a software using the built-in camera. Document images, as scanned with a high resolution flatbed scanner, hardly suffer from irregular illumination, blur, skew and perspective distortion whereas it so happens in case of camera captured document images. Moreover, business cards often have complex background, logo and complex texts like underlined or artistic ones. And thus, neither global nor adaptive binarization can help to isolate the text regions from the card images. On the other hand, mobile devices usually have low computing power (200-666 MHz ARM series processors), less primary memory (upto 128 MB), no Floating Point Unit (FPU) for floating point operations and limited caching. So, methods that involve computationally expensive algorithms and/or high memory requirement, how well be their performance, can not be embedded into the mobile devices for practical applications.

Until recently, various text extraction methods have been proposed and evaluated, of which most of them are for document images. Some have been proposed for business card images captured with a built-in camera of a mobile device [1]-[3]. Few other text extraction methods are reported in [4]-[6]. DCT and Information Pixel Density have been used to analyze different regions of a business card image in [1]. In [2], a low resource consuming region extraction algorithm has been proposed for mobile devices with the limitation that the user needs to manually select the area in which the analysis would be done and the success rate is yet to be improved. Pilu et al. [3] in their work on light weight text image processing for handheld embedded cameras, proposed a text detection method that can not remove the logo(s) of a card and may mistake parts of the oversized fonts as background and can not deal with reverse text. In [4], text lines are extracted from Chinese business card images using document geometrical layout analysis method. Fisher's Discrimination Rate (FDR) based approach followed by various text selection rules is presented in place of mathematical morphology based operations [5]. Yamaguchi et al. [6] has designed a digit extraction method in the works of telephone number identification and recognition from signboards by eliminating noise using Roberts filter, and then applied different text identification rules.

While some of the above methods seem to be computationally expensive, some others need more accuracy. In this paper, we have presented a computationally efficient rule-based text extraction method that works satisfactorily for camera captured business card images under the computing constraints of mobile devices.

## 2. The Present Work

Developed extraction method works mainly in two steps. At the first step, the background of the image is eliminated at a coarse level as discussed in section 2.1. Then, we find and classify the connected components from the background-eliminated card image as discussed in sections 2.2 and 2.3. Fig. 1(a-d) show different operations on a business card image and Fig. 2 shows an enlarged view of some portions of the card image for the different steps in Fig. 1.

### 2.1. Coarse Background Elimination

The entire image is at first divided into blocks of a fixed size. The more the length of the block is, the more the words of the same text line get connected. And similarly, the less the height of the block is, the less is the possibility that a block covers more than one text lines. So, we have mostly experimented with rectangular blocks. The width and height is varied and tuned for best results and we found that it works well for the block of width *W/64* (*W* is the width of the card image) and height 2 pixels. Next, we classify each block as either an information block or a background block based on the intensity variance within it. An information block belongs to either a text region or an image region including noise. The motivation behind this approach is that the intensity variance is low in case of background blocks and high in case of information blocks. So, if the intensity variance of a block is lesser than a dynamic threshold ($T_\sigma$) as given in Eq. (1), it is considered as a background block. Otherwise, the block is considered as an information block. But, no block is classified as background until the minimum intensity within the block exceeds a heuristically chosen threshold ($T_{min}$). The formulation of $T_\sigma$ is described below.

$$T_\sigma = T_{fixed} + T_{var} \quad (1)$$

$$T_{var} = [(G_{min} - T_{min}) - min(T_{fixed}, G_{min} - T_{min})] * 2 \quad (2)$$

where, $G_{min}$ and $G_{max}$ are the minimum and maximum gray level intensity of the pixels in a block respectively and $T_{fixed}$ is the minimum intensity tolerance subject to tuning.

All the pixels of the blocks identified as background in this section are assigned the maximum intensity i.e. 255 to denote that they are part of the background. Fig. 1(b) shows the view after the transformation discussed in this section on Fig. 1(a). Fig. 2(a-c) zoom a portion of Fig. 1(a) and Fig. 1(c-d) respectively.

### 2.2. Identifying Connected Components

After the operation as discussed in Section 2.1, touching blocks together form a region called connected component (CC). A CC is an isolated gray region of the card image as shown in Fig. 1(b). The region growing algorithm [7] has been used to identify the CCs for further processing.

### 2.3. Rule-based classification of the CCs

A connected component may be a picture, logo, noise or a text region. We focus to identify only the text regions using rule-based classification. Whenever we identify a CC, we apply the following rules and decide whether it is a text region or not. We have played a little conservative in the approach so that we do not loose any text region, however unclear it is, even if we classify a non-text region as a text one.

A text region can not be too small because it must contain at least one character. Besides that, small characters are very unlikely to be isolated in the card. So, if the height and width of a CC are lesser than that of the smallest possible character of the card, then the region could be classified as noise.

Let, $H_{cc}$ and $W_{cc}$ be the height and width of a connected component respectively and $A_{cc}$ be its area in terms of pixels. Then, the CC is classified as noise if any of the following three conditions (Eq. (3-5)) gets satisfied. Here, $H_{TH}$, $W_{TH}$ and $A_{TH}$ denote minimum height, width and area of a CC respectively.

$$H_{cc} < H_{TH} \tag{3}$$
$$W_{cc} < W_{TH} \tag{4}$$
$$A_{cc} < A_{TH} \tag{5}$$

Some business cards may contain lines along either or both horizontal and vertical directions. We consider a CC to be a horizontal line if Eq. (6) is satisfied and a vertical line if Eq. (7) is satisfied. $L_{TH}$ and $B_{TH}$ denote the minimum length and maximum breadth of a line.

$$H_{cc} < B_{TH} \ \& \ W_{cc} > L_{TH} \tag{6}$$
$$W_{cc} < B_{TH} \ \& \ H_{cc} > L_{TH} \tag{7}$$

Typically, a text region has a certain range of width to height ratio ($R_{w2h}$). We consider a CC as a text if $R_{w2h}$ lies within the range ($R_{min}$, $R_{max}$).

$$R_{min} < R_{w2h} < R_{max} \tag{8}$$

To remove the logo(s), we have made an assumption that neither a horizontal nor a vertical line can be drawn through a logo and it is not as small as a large possible character within the card. Thus, logos and other noises satisfying the above assumption get eliminated.

Another important property of text region is that in a text region, the number of foreground pixels is significantly less than that of the background pixels. We consider a certain range of ratio of foreground pixels to the background ones ($RA_{cc}$) given by ($RA_{min}$, $RA_{max}$) for candidates of text region.

### 2.4. Binarization

As a CC is classified as a text region, it is binarized with adaptive yet simple technique. If the intensity of a pixel within the CC is less than the mean of the maximum and minimum intensities of a CC, it is taken as a foreground pixel. Otherwise, we check the 8 neighbors of the pixel and if any 5 or more neighbors are foreground, then also we consider the pixel as a foreground one. It may be noted that the border pixels do not have 8 neighbors and so will not be subject to this technique. The remaining pixels are considered as part of the background. The algorithm is given in Table 1.

**Table 1: Binarization Algorithm**

```
for all pixels (x, y) in a CC
    if Intensity(x, y) < (G_min + G_max)/2, then
        mark (x, y) as foreground
    else
        if no. of foreground neighbors > 4, then
            mark (x, y) as foreground
        else
            mark (x, y) as background
        end if
    end if
end for
```

The advantage of this approach of binarization is that the disconnected foreground pixels of a character are likely to become connected due to neighborhood consideration. Instead of having efficient binarization techniques, we have implemented this simple one keeping the computational constraints of the mobile devices in view.

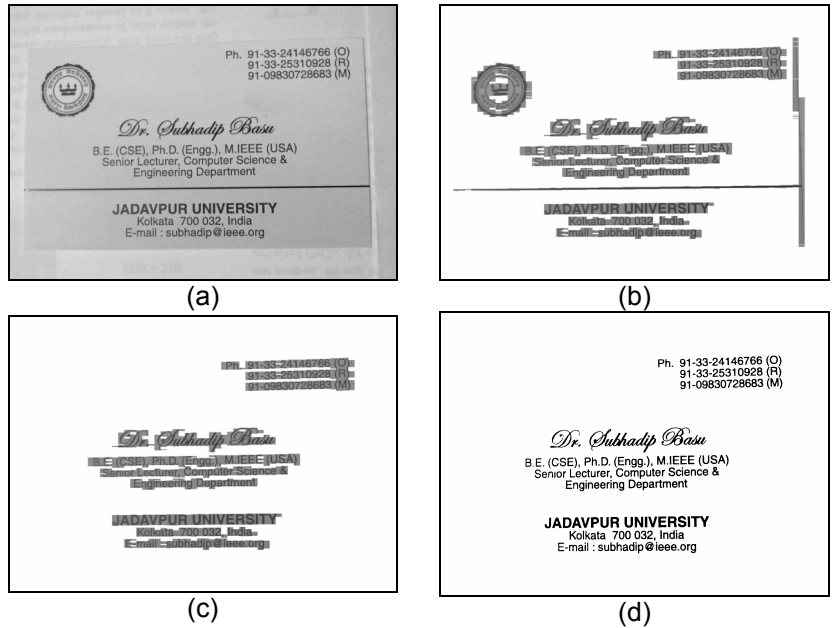

**Fig. 1: Text Extraction from a business card image**
  (a) Original card image
  (b) View after coarse background elimination
  (c) Extracted text regions
  (d) Binarized card image

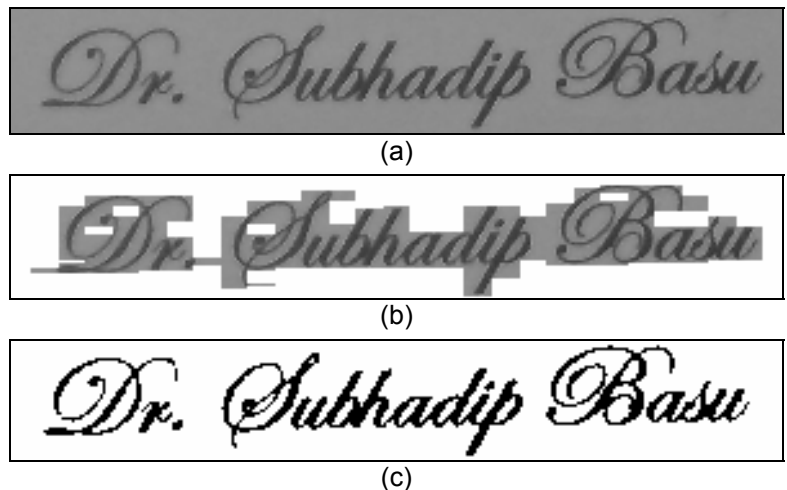

**Fig. 2: Enlarged view of some portions of Fig. 1**
  (a) As in original image
  (b) Single connected component formation after coarse background elimination
  (c) Post-binarization view

## 3. Experimental Results and Discussion

To evaluate the performance of the proposed text region extraction method, we have experimented on a dataset of 100 business card images of various types acquired with a cell-phone camera (Sony Ericsson K810i). The dataset contains both simple and complex cards including complex backgrounds and logos. Some cards contain multiple logos and some logos are combination of text and image. Most of the images are skewed, perspectively distorted and degraded. We process the images without any kind of rectification or enhancement.

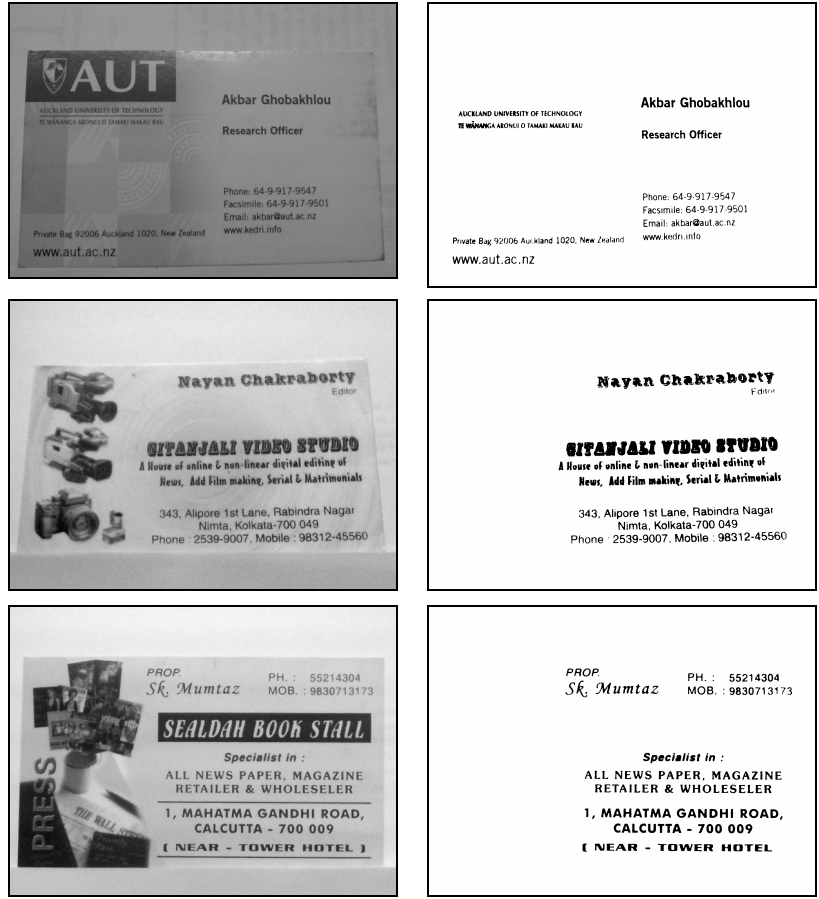

**Fig. 3: Sample cards (left) and their binarized view (right)**

To quantify the text extraction accuracy, we have designed the following method. A text or a background region can be identified either way. Background here refers to all non-text regions. So, there can be four possible cases as shown in Table 2. BB and TT are considered as true classifications whereas BT and TB are false ones.

**Table 2: Justification of CCs classification**

| CC/Region | Classified as | Justification |
|---|---|---|
| Background | Background | BB (True) |
| Background | Text | BT (False) |
| Text | Background | TB (False) |
| Text | Text | TT (True) |

So, the text extraction accuracy of the current technique is defined as

$$\text{Accuracy} = \frac{\text{No. of true classifications}}{\text{Total no. of CCs}} \qquad (9)$$

Following the above text extraction accuracy quantization method, we have got a maximum success rate of 98.93% for 3 Mega pixel images with $T_{fixed}$ = 20, $T_{min}$ = 100, $H_{TH}$ = $H$/60, $W_{TH}$ = $W$/40, $A_{TH}$ = $W*H$/1500, $B_{TH}$ = $H$/100, $L_{TH}$ = $W$/40, $R_{min}$ = 1.2, $R_{max}$ = 32, $RA_{min}$ = 5 and $RA_{max}$ = 90. Table 3 shows the accuracy rates when experimented with other resolutions on a moderately powerful desktop (PIV 2.4 GHz Processor, 256 MB RAM, 1 MB L2 Cache).

An observation [8] reveals that the majority of the processing time of an Optical Character Recognition (OCR) engine embedded into a mobile device is consumed in preprocessing including binarization. Although, we have shown the computational time of the presented method with respect to a desktop, the total time required to run the developed method on mobile devices will be tolerable. Fig. 4 shows the computation time with various resolutions.

**Table 3: Classification Accuracy with Various Resolutions**

| Resolution (width x height) | Mean accuracy (%) |
|---|---|
| 640x480 (0.3 MP) | 97.80 |
| 800x600 (0.45 MP) | 97.86 |
| **1024x768 (0.75 MP)** | **98.54** |
| 1182x886 (1 MP) | 98.00 |
| 1672x1254 (2 MP) | 98.45 |
| 2048x1536 (3 MP) | 98.93 |

As, limited memory is another constraint of the mobile devices, the presented method is designed to work with low memory requirement. Fig. 5 shows the average memory consumption with images of various resolutions.

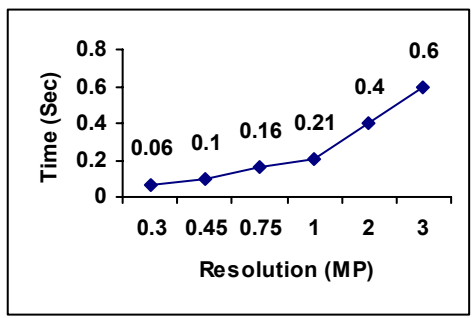
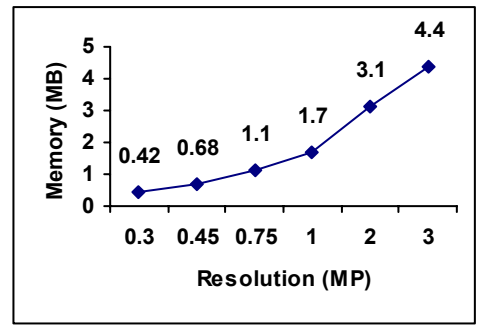

**Fig. 4: Computation Time with Various Resolutions**

**Fig. 5: Memory Consumption with Various Resolutions**

Our aim is to deploy the proposed method into mobile devices. Therefore, we want to develop a light-weight Business Card Reader (BCR) system beforehand and then to embed into the devices because the challenges are known and one can easily have a decent idea about the timing and memory requirement for an architecture once specified.

Although, one can see that the developed method works well as shown in Fig. 3, it too has certain limitations. Sometimes, the dot of 'i' or 'j' gets removed. When text and image/logo are very near to each other, they together form a single CC and get wrongly classified as background or text as shown in Fig 6. The current technique does not work for the white texts on a dark background as shown in the last image of Fig. 3. Our future work will include that.

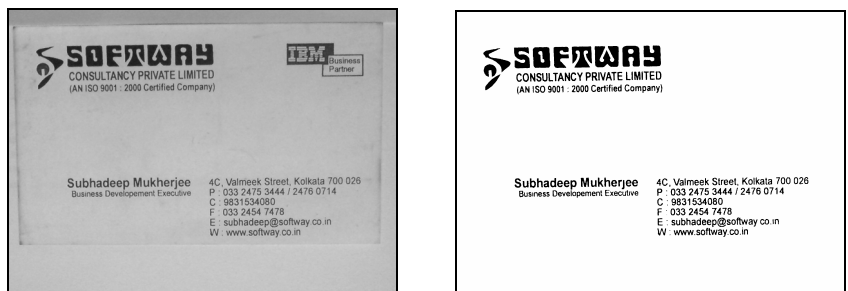

**Fig. 6: Sample card (left) in which the logo is misclassified as text (right)**

## 4. Conclusion

We have presented and evaluated a method of text extraction and binarization for mobile camera captured business card images and our experiment shows that the result is satisfactory. It can be observed from this experimentation, that with the increase in image resolution, the computation time and memory requirements increase proportionately. Although, the maximum accuracy is obtained with 3 mega pixel resolution, it involves high memory requirement and 0.6 seconds of processing time. It is evident from the findings that the optimum performance is achieved at 1024x768 (0.75 MP) pixels resolution with a reasonable accuracy of 98.54% and significantly low (in comparison to 3 MP) processing time of 0.16 seconds and memory requirement of 1.1 MB. As we have employed a little conservative approach and got some background CCs as text ones (BT), we hope that these will get removed in segmentation and recognition. The occurrence of text classified as background (TB) is absolutely less which reveals good potential for the developed method discussed in this paper. Moreover, the presented method is directly applied on the raw images which are heavily distorted. If the images are rectified and then passed to our engine, we hope and foresee a better accuracy.

**Acknowledgement**

Authors are thankful to the *Center for Microprocessor Application for Training Education and Research (CMATER)* and Project on *Storage Retrieval and Understanding of Video for Multimedia (SRUVM)* of the Computer Science and Engineering Department, Jadavpur University for providing infrastructural support for the research work. We are also thankful to the *School of Mobile Computing and Communication (SMCC)* for proving the research fellowship to the first author.